\documentclass[final]{cvpr}

\usepackage{times}
\usepackage{epsfig}
\usepackage{graphicx}
\usepackage{amsmath}
\usepackage{amssymb}

\usepackage[ruled]{algorithm2e}
\usepackage{multirow}
\usepackage{diagbox}
\usepackage{placeins}
\usepackage{textcmds}

\newcommand{\algoref}[1]{Algorithm~\ref{#1}}

\usepackage[pagebackref=true,breaklinks=true,colorlinks,bookmarks=false]{hyperref}

\def\cvprPaperID{1578} 
\def\confYear{CVPR 2021}

\begin{document}

\title{Restore from Restored: Single Image Denoising with Pseudo Clean Image}

\author{Seunghwan Lee\textsuperscript{1}, Dongkyu Lee\textsuperscript{1,2}, Donghyeon Cho\textsuperscript{3}, Jiwon Kim\textsuperscript{4}, and Tae Hyun Kim\textsuperscript{1}\\

\textsuperscript{1}Dept. of Computer Science, Hanyang University, Seoul, Korea\\
{\small\textit{\{seunghwanlee, leedk\}@hanyang.ac.kr, lliger9@gmail.com}}\\

\textsuperscript{2}NAVER LABS\\

\textsuperscript{3}Dept. of Electronic Engineering, Chungnam National University, Daejeon, Korea\\
{\small\textit{cdh12242@gmail.com}}\\

\textsuperscript{4}SK T-Brain, Seoul, Korea\\
{\small\textit{jk@sktbrain.com}}\\
}


\maketitle

\begin{abstract}
In this study, we propose a simple and effective fine-tuning algorithm called \qq{restore-from-restored}, which can greatly enhance the performance of fully pre-trained image denoising networks.
Many supervised denoising approaches can produce satisfactory results using large external training datasets. However, these methods have limitations in using internal information available in a given test image.
By contrast, recent self-supervised approaches can remove noise in the input image by utilizing information from the specific test input. However, such methods show relatively lower performance on known noise types such as Gaussian noise compared to supervised methods.
Thus, to combine external and internal information, we fine-tune the fully pre-trained denoiser using pseudo training set at test time. By exploiting internal self-similar patches (i.e., patch-recurrence), the baseline network can be adapted to the given specific input image.
We demonstrate that our method can be easily employed on top of the state-of-the-art denoising networks and further improve the performance on numerous denoising benchmark datasets including real noisy images.
\end{abstract}

\section{Introduction}

\iffalse
When a scene is captured by an imaging device, a desired clean image $\mathbf{X}$ is corrupted by noise $\mathbf{n}$. We usually assume that the noise $\mathbf{n}$ is an Additive White Gaussian Noise (AWGN), and the observed noisy image $\mathbf{Y}$ can be expressed as $\mathbf{Y} = \mathbf{X} + \mathbf{n}$.
In particular, noise $\mathbf{n}$ increases in environments with high ISO, short exposure times, and low-light conditions.
Image denoising is a task that restores the clean image $\mathbf{X}$ by removing the added noise $\mathbf{n}$ from the input $\mathbf{Y}$, and is a highly ill-posed problem. Thus, substantial literature concerning denoising problem has been introduced~\cite{wavelet_denoise,NLM,dictionary_denoise,Nuclear_denoise,BM3D,nonlocal_sparse_denoise,centralized_sparse_denoise,color_sparse_denoise}.
\else
When a scene is captured by an imaging device, a desired clean image can be corrupted by noise and the amount of added noise increases in environments with high ISO, short exposure time, and low-light condition.
Image denoising is a task that restores the clean image by removing the noise in the noisy image and is a highly ill-posed problem. Accordingly, substantial literature concerning denoising problem has been studied~\cite{wavelet_denoise,NLM,dictionary_denoise,BM3D,color_sparse_denoise,nonlocal_sparse_denoise,centralized_sparse_denoise,Nuclear_denoise}.
\fi

Recent deep learning technologies have been used to restore a clean image by exploiting large external datasets. These methods train networks for denoising in a supervised manner by using pairs of noisy and ground-truth clean images and have shown satisfactory results.
However, the supervised methods are limited in performance when the noise distribution of the test image is considerably different from that of noise in the training dataset (\ie, when domain misalignment occurs). To overcome this problem, researchers have proposed self-supervised training methods, such as noise-to-noise~\cite{Noise2noise}, noise-to-void~\cite{Noise2void}, and noise-to-self~\cite{Noise2self}, which allow the networks to learn the noise of the input image with unknown distribution at the test stage without using the ground-truth clean image. However, the performance of the self-supervised approaches is inferior compared with the supervised methods when the noise distribution of the test image is known in advance (\eg., Gaussian noise) due to the lack of generalization power learnable from a large external database.

In this work, we aim to remove known and unknown noise by combining supervised and self-supervised approaches and elevate the performance of existing denoising networks through a method that updates the network parameters by using the information available from the given noisy input image at the test time.
Specifically, we start with a fully pre-trained denoising network trained either in a supervised or self-supervised manner.
Then, the network is adapted (fine-tuned) by using the proposed algorithm that can overcome the limitations of conventional supervised methods~\cite{RIDNet,DnCNN} and self-supervised approaches~\cite{Noise2void,Noise2self}.
In particular, we greatly improve the denoising performance by exploiting self-similarity in the noisy input image; self-similarity is a property wherein a large number of similar patches are recurring within a single image and has been studied in numerous denoising and super-resolution tasks to enhance the image quality~\cite{BM3D,SADCT, glasner,zssr,selfex}.
Moreover, our method does not rely on a certain architecture of networks and can be easily integrated with conventional denoising networks without any modification.

In this study, we present a simple yet effective network adaptation (fine-tuning) algorithm that allows us to upgrade the denoising networks by combining supervision and self-supervision. The contributions of this paper are summarized as follows:
\begin{itemize}
	\item We demonstrate that conventional denoising networks can be further improved using self-similar patches available from the input test image.
	\item We present a new learning algorithm that effectively combines supervised and self-supervised methods and achieve the state-of-the-art denoising performance on numerous benchmark datasets including not only synthetic noise but also real noise.
	\item Our algorithm can be easily embedded on top of conventional denoising networks without modifying the original network architectures.
\end{itemize}

\section{Related Work}
In this section, we review numerous denoising methods with and without using the ground-truth clean images for training. 

Image denoising is an actively studied area in image processing, and various denoising methods have been introduced, such as self-similarity-based methods~\cite{NLM,BM3D,SADCT}, sparse-representation-based methods~\cite{nonlocal_sparse_denoise,Sparse11}, and external database exploiting methods~\cite{Adaptive15,External15,Category_specific,External_category}. With the recent development of deep learning technologies, denoising technique has been also improved, and remarkable progress has been achieved in this field. Specifically, after Xie~\etal~\cite{Denoising12} adopted deep neural networks for denoising, numerous follow-up studies have been proposed~\cite{DnCNN,IrCNN,FFDNet,RDN,Nonlocal_color,Nonlocal_recurrent,Nonlocal_residual,CBDNet,RIDNet}.

Zhang~\etal~\cite{DnCNN} proposed a deep network on the basis of the deep convolutional neural network (CNN) to learn a residual image with a skip connection between the input and the output of the network, accelerated the training speed, and enhanced the denoising quality.
Zhang~\etal~\cite{IrCNN} also proposed IRCNN. This network can be combined with conventional model-based optimization methods to solve various image restoration problems, such as denoising, super-resolution, and deblurring. 
Furthermore, Zhang~\etal~\cite{FFDNet} proposed a fast and efficient denoising network FFDNet, which takes cropped sub-images and noise level as inputs. 
In addition to being fast, FFDNet can handle locally varying and a wide range of noise levels.
Zhang~\etal~\cite{RDN} introduced a deep residual dense network (RDN), which is composed of multiple residual dense blocks. RDN achieves superior performance by exploiting both the local and global features through densely connected convolutional layers and dense feature fusion. Zhang~\etal~\cite{Nonlocal_residual} proposed a residual non-local attention network (RNAN), which consists of a trunk and (non-) local mask branches, to capture long-range dependencies among pixels. The non-local block was used with a recurrent mechanism to increase the receptive field of the denoising network~\cite{Nonlocal_recurrent}. Recently, CBDNet~\cite{CBDNet} and RIDNet~\cite{RIDNet} were introduced to handle real noise where the noise distribution is unknown (blind denoising). CBDNet is a two-step approach that combines noise estimation and non-blind denoising tasks, whereas RIDNet is a single-stage method that employs feature attention in the middle of the network. More recently, MIRNet~\cite{MIRNet} achieved state-of-the-art results for a variety of image restoration tasks including image denoising. MIRNet proposed parallel multi-scale convolution streams to extract multi-resolution features for both spatial details and contextual information. MIRNet also proposed attention-based feature aggregation to fuse features across all the scales for improved feature representations.

After deep CNN was adopted to increase denoising performance, various research directions, such as residual learning for constructing deep networks, non-local or hierarchical features for enlarging the receptive fields, and noise estimation for handling real photographs, have been considered. However, such works still remain restricted to the cases in which networks are supervised by target clean images. To overcome this problem, several self-supervision-based studies have been conducted to train the network without using the ground-truth clean images.
Lehtinen~\etal~\cite{Noise2noise} introduced a \emph{noise-to-noise} approach and demonstrated that a denoising network can be trained without using clean images. The denoising network was trained with pairs of differently corrupted patches on the basis of statistical reasoning that the expectation of randomly corrupted signal is close to the clean signal. Krull~\etal~\cite{Noise2void} proposed a self-supervised \emph{noise-to-void} method to deal with noise with unknown distribution in the given input image.
Baston and Royer~\cite{Noise2self} also introduced a \emph{noise-to-self} method to train the network without knowing the ground-truth data at the test time. 
However, these self-supervision-based methods cannot outperform supervised methods when the noise distribution of the input image is known.
Thus, we aim to benefit from the supervised and self-supervised approaches and develop a new algorithm that can handle not only synthetic but also real noisy images by integrating the external and internal information.




\begin{figure*}[h]
\begin{center}
\includegraphics[width=0.7\linewidth]{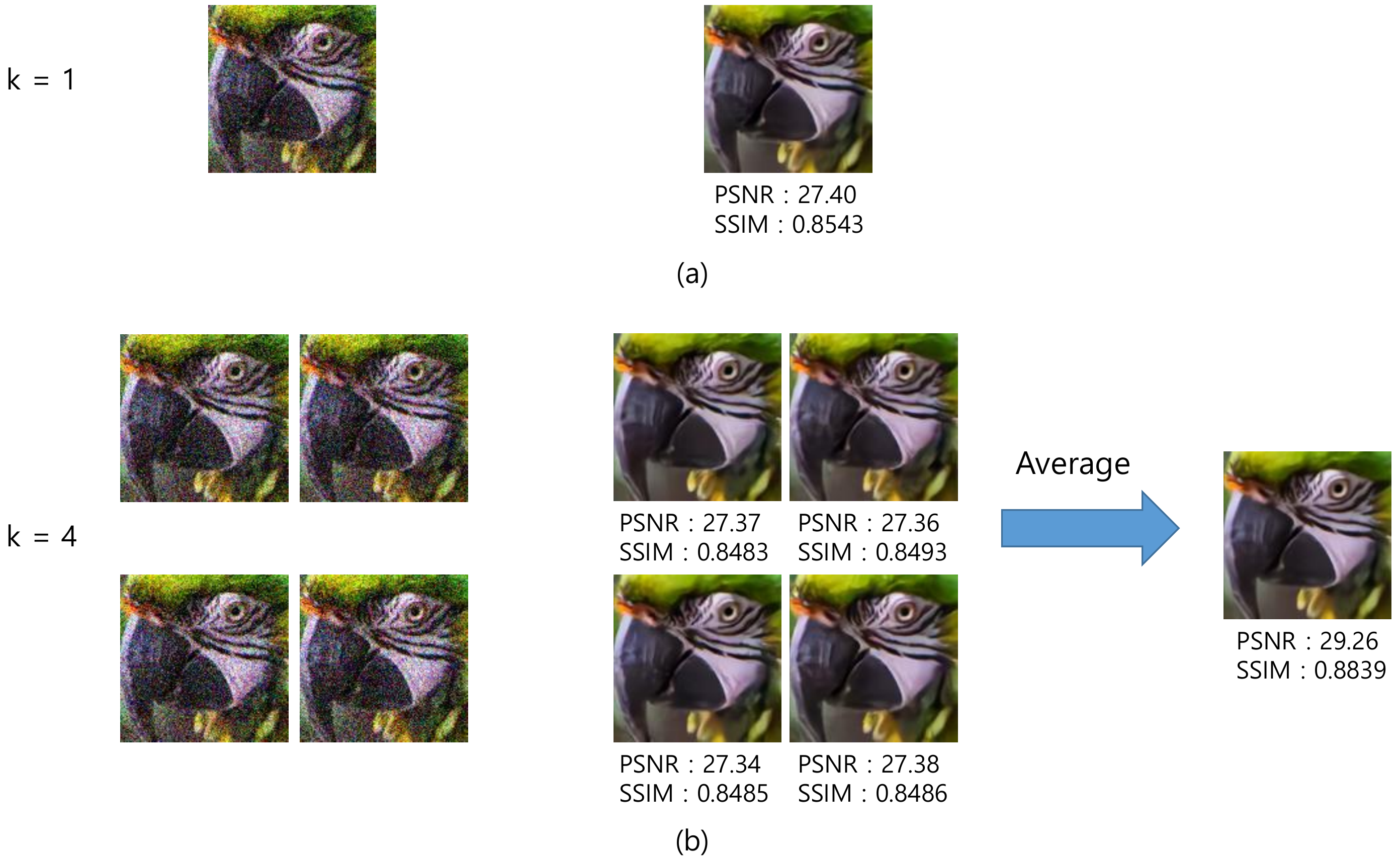}
\end{center}
\caption{ (a) \textbf{Left to right}: a noisy input patch with Gaussian random noise (i.e., $k=1$) and denoising result by DnCNN~\cite{DnCNN}.
(b) \textbf{Left to right}: four differently corrupted patches with Gaussian random noise (i.e., $k=4$), denoising results by DnCNN~\cite{DnCNN}, and averaged version of the denoised patches. Denoising quality increases by taking an average of four recurring patches.}
\label{patch_recurrence_averaging}
\end{figure*}

\section{Denoising with Self-similarity}

\begin{table*}[h]
\centering
\begin{tabular}{|c|c|c|c|c|c|}
\hline
k (number of  self-similar patches) & 1           & 4           & 9           & 16          & 25          \\ \hline
DnCNN~\cite{DnCNN}  & 27.40/0.8543 & 29.26/0.8839 & 29.71/0.8937 & 29.83/0.8961 & \textbf{29.89/0.8971} \\ \hline
RDN~\cite{RDN}    & 27.57/0.8524 & 29.35/0.8861 & 29.76/0.8932 & 29.93/0.8963 & \textbf{29.98/0.8968} \\ \hline
RIDNet~\cite{RIDNet} & 27.66/0.8558 & 29.52/0.8901 & 29.94/0.8960 & 30.10/0.8981 & \textbf{30.18/0.9001} \\ \hline
N2S~\cite{Noise2self} & 24.37/0.7000 & 25.37/0.7629 & 25.48/0.7722 & \textbf{25.49}/0.7762 & 25.48/\textbf{0.7778} \\ \hline
\end{tabular}
\vspace{1Ex}
\caption{Results by averaging $k$ denoised self-similar patches. 
For denoising, DnCNN~\cite{DnCNN}, RDN~\cite{RDN}, RIDNet~\cite{RIDNet}, and N2S~\cite{Noise2self} are used. 
Note that DnCNN, RDN, and RIDNet are supervised methods where pre-trained parameters are available, but, N2S~\cite{Noise2self} is a self-supervised one which requires test-time training.}
\label{table_patch_recurrence}
\end{table*}

Using multiple self-similar patches within a test input image is a key property that leads to success of our method. In this section, we present a toy example as illustrated in~\figref{patch_recurrence_averaging} and demonstrate how the self-similarity can improve the performance of conventional supervised and self-supervised denoising methods.

In~\figref{patch_recurrence_averaging}, 
we first generate a clean image $\textbf{X}_k$ which includes $k$ identical (self-similar) patches.
Next, we synthesize a noisy image $\textbf{Y}_k$ by adding zero-mean Gaussian random noise to $\textbf{X}_k$.
Then, the synthetic noisy image $\textbf{Y}_k$ can be passed into any conventional supervised/self-supervised denoising network as input and we can obtain a denoised image $\tilde{\textbf{X}}_k$ as an output. Finally, we compute an averaged version of the $k$ denoised patches in the denoised image $\tilde{\textbf{X}}_k$. We observe that the amount of noise can be further reduced (see the PSNR/SSIM changes) by simply taking an average of $k$ self-similar denoised patches since the remaining noise in the denoised patch is different from each other. Specifically, in Table~\ref{table_patch_recurrence}, PSNR and SSIM values of the averaged patches from various supervised and self-supervised denoising results are provided. We see that the denoising quality improves consistently as the number of self-similar patches used for averaging increases (\ie, large $k$). Please refer to our supplementary material for more results.

Based on this observation, we aim to develop a new learning algorithm, which can efficiently estimate this averaged version of similar patches scattered within a given denoised image without explicit searching and aligning the self-similar patches. Moreover, we provide a thorough analysis and explain the mechanism of the proposed algorithm in the following sections.

\section{Proposed Method}

Supervised denoising networks can explore the information available from the large external datasets, but they cannot utilize internal information (\ie, self-similarity) within a given test input image. Thus, specialized network modules, such as non-local operator to search self-similar patches/features within the input, are introduced to mitigate this problem~\cite{Nonlocal_color,Nonlocal_recurrent}. Nevertheless, these techniques still restrict the search range of the non-local operator due to the limited capacity of the networks (e.g., memory). By contrast, self-supervised approaches, such as noise-to-void~\cite{Noise2void} and noise-to-self~\cite{Noise2self}, can exploit self-similarity to train the network at test time without this restriction. However, training algorithms of these self-supervised approaches are tricky and it takes a large amount of time because only a small number of pixel locations are considered for back-propagation at each training iteration. Specifically, we find that only about $0.3\%$ and $6.25\%$ of the pixel values are used to train noise-to-void and noise-to-self at each iteration in their official codes.
That is, current test-time training algorithm with self-supervision is inefficient and time-consuming. 
In addition to this, networks trained with self-supervision cannot surpass the networks trained with supervision when the noise distribution of the input image is known.
Thus, we present a new training algorithm that combines supervised and self-supervised approaches and takes benefits from these two different learning algorithms.

\begin{figure*}[h]
\begin{center}
\includegraphics[width=0.8\linewidth]{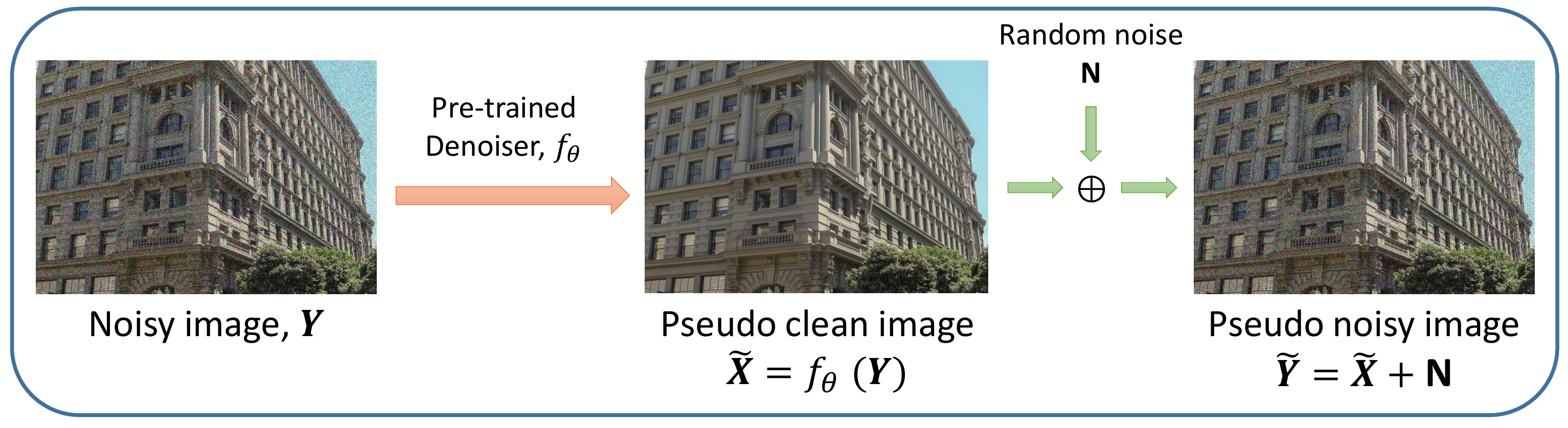}
\end{center}
\caption{Our dataset generation for the self-supervised learning.}
\label{pseudo_generation}
\end{figure*}

\begin{figure*}[h]
\begin{center}
\includegraphics[width=1.0\linewidth]{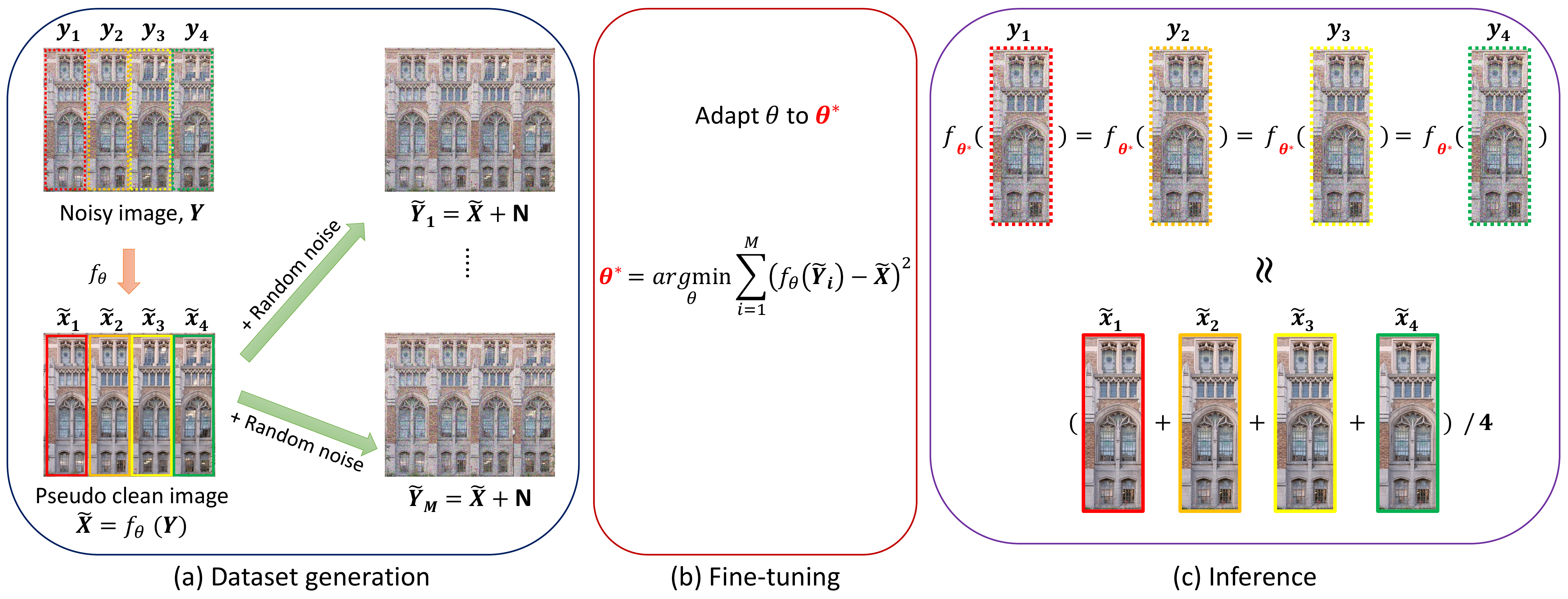}
\end{center}
\caption{(a) Four dashed patches ($\mathbf{y}_j$) in the input $\mathbf{Y}$ and solid ones ($\tilde{\mathbf{x}}_j$) in the pseudo clean image $\tilde{\mathbf{X}}$ denote self-similar patches. We generate pseudo noisy images $\{\Tilde{\textbf{Y}}_1,..., \Tilde{\textbf{Y}}_M\}$ by adding random noise to $\tilde{\mathbf{X}}$. 
(b) We adapt (fine-tune) the pre-trained parameter $\theta$ to $\theta^*$ to the given specific input using the acquired dataset.
(c) During the inference, we obtain the averaged version of self-similar patches (=$\frac{(\tilde{\mathbf{x}}_1+ \tilde{\mathbf{x}}_2+\tilde{\mathbf{x}}_3+\tilde{\mathbf{x}}_4)}{4})$
as the output of $\mathbf{f}_{\theta^*}(\mathbf{y}_j)$ for any $j \in \{1,2,3,4\}$.}
\label{fig_rfr_steps}
\end{figure*}

\subsection{Restore-from-Restored Method}
In this work, we introduce a simple yet effective denoising algorithm that can enhance the denoising results obtained from conventional methods by adapting the pre-trained network parameters to the specific input image given at test time without modifying their original architectures.

Many supervised approaches~\cite{DnCNN, RIDNet} employ clean images in the large external dataset as target images to train the networks and self-supervised methods~\cite{Noise2void, Noise2self} use the noisy input image itself as a target for the test-time training.
Compared with conventional training algorithm, our proposed method uses the denoising result predicted by the pre-trained denoising networks as training target during the fine-tuning. Note that the initial network can be pre-trained in a supervised or self-supervised manner. 

The loss function for updating the network parameter during the fine-tuning stage is based on MSE criterion and it is given by,
\iffalse
\begin{equation}
Loss(\theta) = \sum_{i=1}^N \sum_{k=1}^K (\textbf{f}_{\theta}(\textbf{T}(\tilde{\textbf{x}}_k)+\mathbf{n}_{i,k})-\textbf{T}(\tilde{\textbf{x}}_k))^2,
\label{equ_proposed_patch_loss}
\end{equation}

\else

\begin{equation}
Loss(\theta) =  \mathbf{E}[(\textbf{f}_{\theta}(\tilde{\textbf{X}}+\textbf{N})-\tilde{\textbf{X}})^2],
\label{equ_proposed_loss}
\end{equation}
\fi
where $\textbf{f}$ is the denoising network and $\theta$ denotes the network parameter.
Our loss function includes the initially denoised result $\tilde{\mathbf{X}}$ from the fully-trained conventional denoising networks as the training target and its noisy version $\tilde{\mathbf{X}} + \mathbf{N}$ as the training input.
Random noise $\mathbf{N}$ is realized by the mechanism used to pre-train the baseline network (please refer to Sec.~\ref{label_experiments} for details) and the dataset generation mechanism for the fine-tuning is illustrated in~\figref{pseudo_generation}.
Note that our training target $\tilde{\mathbf{X}}$ for the fine-tuning is a \emph{pseudo clean image} rather than a ground-truth clean image or noisy image itself compared with conventional supervised/self-supervised approaches.
Moreover, we can augment the training set by using random transformations (\eg., flip and flop). Augmentations allow our solution to avoid a trivial network parameter $\theta$ that always outputs $\tilde{\mathbf{X}}$ regardless of the network input. We omit the random transformations for the augmentation in this study for the sake of notational simplicity.

To fine-tune the network parameter $\theta$, we assume that the distributions of our train input $\tilde{\mathbf{X}} + \mathbf{N}$ and the real noisy input $\mathbf{Y}$ are similar. Under this assumption, we can learn a desired parameter $\theta^*$ by optimizing the proposed loss function in (\ref{equ_proposed_loss}) and further improve the denoising performance of the baseline network. Moreover, we can also use L0 or L1 loss in the proposed loss function depending on the noise characteristic of the given input image as suggested in~\cite{Noise2noise,frame-to-frame,Noise2void,Noise2self}.

The proposed method is dubbed \qq{\textit{restore-from-restored}} because our denoising model learns parameters by using the initially restored image $\tilde{\mathbf{X}}$, which is the pseudo clean target.

\paragraph{Overall flow} 
Overall sketch of the proposed \qq{\textit{restore-from-restored}} algorithm is described in~\algoref{algorithm_proposed}. 

First, We generate a pseudo clean image $\tilde{\mathbf{X}}$ by removing noise from the given test image $\mathbf{Y}$ by using a pre-trained denoiser. Second, we generate training set for fine-tuning with pairs of pseudo clean image and its corrupted versions with the random noise. Next, we update the network parameter by using the acquired training set. We can use conventional optimization methods, such as SGD and ADAM, for the update. We fine-tune the parameter several times (=$M$) and obtain the final clean image $\textbf{f}_{\theta^*}(\textbf{Y})$ by feeding the noisy input $\mathbf{Y}$ to the fined-tuned denoising network as input.

\begin{algorithm}[h]
	\textbf{Input:} input noisy image $\textbf{Y}$
	
	\textbf{Output:} denoised image $\textbf{f}_{\theta^*}(\textbf{Y})$
	
	\textbf{Require:} denoising network $\textbf{f}$ and the pre-trained parameter $\theta$, pseudo clean image $\tilde{\mathbf{X}}$, the number of iterations $M$, learning rate $\alpha$\\

	\nl i $\leftarrow$ 0
	
	\While{i $<$ M}{
	    \tcp{Loss computation. Random transformation can be applied.}
		\nl \textit{Loss}($\theta$) = $(\textbf{f}_{\theta}(\tilde{\textbf{X}} + \textbf{N})- \tilde{\textbf{X}})^2$

		\nl
		
		\tcp{Parameter update}
		\nl $\theta \leftarrow \theta - \alpha \nabla_{\theta}  \textit{Loss}(\theta)$

		\nl
		
		\nl i $\leftarrow $ i+1  
	}

	\nl $\theta^* \leftarrow \theta$

	\textbf{Return:}  $\textbf{f}_{\theta^*}(\textbf{Y})$
	\caption{\textit{Restore-from-restored}}
	
	\label{algorithm_proposed}
\end{algorithm}

\paragraph{Training principle}

In \figref{fig_rfr_steps}, we provide an illustration to help understanding the principle of our training algorithm.
Given the $k$ self-similar patches $\{\tilde{\mathbf{x}}_1,\dots,\tilde{\mathbf{x}}_k \}$ within a pseudo clean image $\tilde{\mathbf{X}}$ (patches with solid bounding boxes in \figref{fig_rfr_steps}(a)), we can synthesize multiple noisy images by adding random noise $\mathbf{N}$ to $\tilde{\mathbf{X}}$. 
If the baseline denoiser is a fully convolutional network (FCN), due to the translation equivariant nature of the FCN~\cite{cohen2016group},
our loss function becomes:
\begin{equation}
\begin{split}
    Loss(\theta) = &~\mathbf{E}[ 
     ( \mathbf{f}_{\theta}(\tilde{\mathbf{X}} + \mathbf{N}) - \tilde{\mathbf{X}} )^2]\\
    \approx &~\sum_{j=1}^k \mathbf{E}[ 
      (\mathbf{f}_{\theta}(\tilde{\mathbf{x}}_j + \mathbf{n}_j) - \tilde{\mathbf{x}}_j )^2],
\end{split}
\end{equation}
where $\tilde{\mathbf{x}}_j + \mathbf{n}_j$ denotes a self-similar noisy patch in $\tilde{\mathbf{X}} + \mathbf{N}$.
Assuming that a real noisy patch $\mathbf{y}_j$ and the synthetic noisy patch $\tilde{\mathbf{x}}_j + \mathbf{n}_j$ have similar distributions, we can estimate the final denoised patches with the fine-tuned parameter $\theta^*$ as follows:
\begin{equation}
    \mathbf{f}_{\theta^*} (\mathbf{y}_1)=\mathbf{f}_{\theta^*} (\mathbf{y}_2)=\cdots=\mathbf{f}_{\theta^*} (\mathbf{y}_k) \approx \frac{1}{k}\sum_{j=1}^k \tilde{\mathbf{x}}_j,
\label{equ_denoised_patch}
\end{equation}
which is an averaged version of the denoised $k$ self-similar patches.
Not that, the noise variance of the averaged result in (\ref{equ_denoised_patch}) is further reduced by a factor of $k$ compared with that of the pseudo clean image $\tilde{\mathbf{X}}$.
In general, the remaining noise in $\tilde{\mathbf{X}}$ is smaller than the noise in the input image $\mathbf{Y}$ and ours achieves much smaller results.
Moreover, in the proposed loss function, we do not need to explicitly search and align (register) self-similar patches to leverage them only if the baseline denoiser is an FCN. 
Indeed, the number, size and shape of these patches are not restricted in the real scenario and our algorithm can be applicable to any natural image.

\paragraph{Spatially varying denoising performance}
In (\ref{equ_denoised_patch}), our denoising performance is depending on the number of self-similar patches (\ie, $k$) within the given input image. Therefore, we can expect non-uniform and spatially varying denoising performance when we remove noise in an input noisy image $\textbf{Y}$. Specifically, regions with highly recurring patches can be restored better than those with low-recurrence-rate. Thus, rich self-similarity in the given image is the key in the proposed denoising algorithm.

\subsection{Noise-to-Void (Noise-to-Self) vs. Restore-from-Restored}

Given that $\tilde{\textbf{X}}$ is a pseudo clean image, we can fine-tune the network without using the ground-truth image. Our training dataset for fine-tuning are pairs of initially denoised image $\tilde{\textbf{X}}$ and its newly corrupted versions with random noise $\mathbf{N}$. While noise-to-void~\cite{Noise2void} and noise-to-self~\cite{Noise2self} use the noisy input image $\textbf{Y}$ itself as the train target rather than $\tilde{\textbf{X}}$.

Minimizing the proposed loss function in (\ref{equ_proposed_loss}) with the pseudo clean target has several benefits compared with optimizing the loss functions used in noise-to-void and noise-to-self as in the following.

\begin{itemize}
\item Noise-to-void and noise-to-self employ a masking scheme that randomly selects pixel locations for back-propagation; thus, only a small number of pixels are used at each train iteration, thereby making it inefficient. By contrast, our method uses every pixel value for computing gradients and facilitates efficient training.
\item Our loss function depends on the pseudo clean image $\tilde{\textbf{X}}$ rendered from the pre-trained network. Thus, we can not only exploit the internal information (\ie, self-similarity) but also explore the external dataset used for pre-training. In contrast, noise-to-void and noise-to-self restore the averaged version of self-similar \emph{noisy} patches. Meanwhile, our method achieves the averaged version of self-similar \emph{denoised} patches which includes much less noise.

\item Our fine-tuning algorithm is applicable to any state-of-the-art denoising networks only if they are fully convolutional even they are very deep and large. The fined-tuned denoisers can further elevate the denoising performance than that of the baseline denoising networks.

\end{itemize}

\section{Experimental Results}\label{label_experiments}

\subsection{Implementation details}
In our experiments, we evaluate the proposed training algorithm with numerous supervised/self-supervised denoising methods as our baseline networks, such as DnCNN~\cite{DnCNN}, RDN~\cite{RDN}, RIDNet~\cite{RIDNet}, CBDNet~\cite{CBDNet}, MIRNet~\cite{MIRNet} and N2S~\cite{Noise2self}. Our source codes will be publicly available upon acceptance.

First, to remove random Gaussian noise, we pre-train DnCNN, RDN, and RIDNet with Gaussian random noise ($\sigma$ is randomly chosen from [0, 50]) on the DIV2K training set. During the pre-training phase, we minimize the L1 loss between the ground-truth clean images and the network outputs with Adam (the learning rate started from 1e-4 and gradually decreased to 1e-6), and conventional data augmentation techniques are applied (\eg., flip, flop, and rotation). 
Currently, RDN shows the state-of-the-art results in removing Gaussian noise, and RIDNet provides competitive results. We use the light version of RDN (D = 10, C = 4, and G = 16; refer to~\cite{RDN} for details) due to the limited capacity of our graphics unit.

Next, in the real noise removal, we use RIDNet~\cite{RIDNet}, CBDNet~\cite{CBDNet}, MIRNet~\cite{MIRNet} and N2S~\cite{Noise2self} as our baseline networks. We use the fully pre-trained parameters officially provided for RIDNet and MIRNet, and employ the trained parameter provided by a third party for CBDNet.
\footnote{Pytorch version. \url{https://github.com/IDKiro/CBDNet-pytorch}}
To fine-tune the network parameters for real noise removal,  we use the noise model introduced in CBDNet~\cite{CBDNet} to generate our pseudo noisy images (i.e., $\tilde{\mathbf{X}}+\mathbf{N}$ in (\ref{equ_proposed_loss})) during the fine-tuning stage and optimize the parameters by using ADAM (learning rate = 1e-5).
Specifically, the real noise model used in CBDNet~\cite{CBDNet} considers an in-camera image signal processing (ISP) pipeline. First, a camera response function (CRF) is randomly chosen from 201 different CRFs which are provided in CBDNet, at each fine-tuning iteration. Then, heteroscedastic Gaussian model (\ie, Poissonian-Gaussian noise model) is applied to the raw image space. We evaluate the performance of the proposed fine-tuning algorithm in removing real noise on the public real noise benchmark datasets, such as SIDD+~\cite{SIDDplus} and Nam~\cite{Nam}.

\begin{table*}[h]
\centering
\resizebox{\textwidth}{!}{%
\renewcommand{\arraystretch}{1.3}
{\huge
\begin{tabular}{|c|c|c|c|c|c|c|c|c|c|c|}
\hline
                        & Dataset           & \multicolumn{3}{c|}{Urban100} & \multicolumn{3}{c|}{DIV2K} & \multicolumn{3}{c|}{BSD68} \\ \hline
Method & \backslashbox{Fine-tuning}{Noise level} & $\sigma=$15    & $\sigma=$25    & $\sigma=$40    & $\sigma=$15    & $\sigma=$25    & $\sigma=$40    & $\sigma=$15    & $\sigma=$25    & $\sigma=$40    \\ \hline
\multirow{5}{*}{DnCNN}  & Fully pre-trained & 33.88/0.9413    & 31.40/0.9105    & 29.07/0.8686   & 34.60/0.9413   & 31.99/0.9057   & 29.66/0.8583  & 33.89/0.9293   & 31.21/0.8824   & 28.92/0.8218  \\ \cline{2-11} 
       & $M=10$             & 33.99/0.9418 & 31.54/0.9116 & 29.23/0.8708 & 34.65/0.9416 & 32.04/0.9063 & 29.73/0.8595 & 33.95/0.9299 & 31.27/0.8832 & 28.98/0.8237 \\ \cline{2-11} 
       & $M=20$             & 34.04/0.9422 & 31.60/0.9125 & 29.30/0.8723 & 34.67/0.9419 & 32.07/0.9068 & 29.75/0.8601 & 33.97/0.9300 & 31.29/0.8838 & 29.01/0.8242 \\ \cline{2-11} 
       & $M=40$             & \textbf{34.09/0.9426} & \textbf{31.65/0.9132} & \textbf{29.36/0.8734} & \textbf{34.70/0.9423} & \textbf{32.09/0.9073} & \textbf{29.77/0.8607} & \textbf{33.98/0.9302} & \textbf{31.31/0.8840} & \textbf{29.02/0.8244} \\ \hline
\multirow{5}{*}{RDN}    & Fully pre-trained & 34.06/0.9428    & 31.64/0.9139    & 29.36/0.8748   & 34.75/0.9429   & 32.15/0.9087   & 29.84/0.8630  & 33.96/0.9303   & 31.29/0.8844   & 29.02/0.8244  \\ \cline{2-11} 
       & $M=10$             & 34.13/0.9431 & 31.73/0.9146 & 29.48/0.8766 & 34.78/0.9434 & 32.19/0.9094 & 29.89/0.8643 & 34.00/0.9308 & 31.34/0.8851 & 29.07/0.8268 \\ \cline{2-11} 
       & $M=20$             & 34.17/0.9434 & 31.78/0.9151 & 29.54/0.8780 & 34.80/0.9435 & 32.21/0.9095 & 29.91/0.8647 & 34.02/0.9307 & 31.35/0.8850 & 29.10/0.8273 \\ \cline{2-11} 
       & $M=40$             & \textbf{34.22/0.9438} & \textbf{31.85/0.9160} & \textbf{29.62/0.8795} & \textbf{34.83/0.9438} & \textbf{32.24/0.9100} & \textbf{29.94/0.8653} & \textbf{34.04/0.9309} & \textbf{31.38/0.8854} & \textbf{29.12/0.8276} \\ \hline
\multirow{5}{*}{RIDNet} & Fully pre-trained & 34.30/0.9454   & 31.94/0.9184   & 29.71/0.8818   & 34.88/0.9441   & 32.29/0.9107   & 29.99/0.8662  & 34.03/0.9308   & 31.37/0.8853   & 29.11/0.8267  \\ \cline{2-11} 
       & $M=10$             & 34.41/0.9463 & 32.06/0.9202 & 29.87/0.8855 & 34.95/0.9453 & 32.36/0.9125 & 30.07/0.8693 & 34.10/0.9319 & 31.44/0.8875 & 29.18/0.8307 \\ \cline{2-11} 
       & $M=20$             & 34.44/0.9466 & 32.11/0.9206 & 29.92/0.8863 & 34.97/0.9455 & 32.38/0.9128 & 30.10/0.8697 & 34.11/0.9320 & 31.46/0.8877 & 29.21/0.8311 \\ \cline{2-11} 
       & $M=40$             & \textbf{34.48/0.9467} & \textbf{32.16/0.9211} & \textbf{29.98/0.8871} & \textbf{34.99/0.9456} & \textbf{32.40/0.9130} & \textbf{30.12/0.8701} & \textbf{34.12/0.9322} & \textbf{31.47/0.8879} & \textbf{29.22/0.8317} \\ \hline
\end{tabular}%
}
}
\caption{Fine-tuning results of state-of-the-art methods (DnCNN~\cite{DnCNN}, RDN~\cite{RDN}, and RIDNet~\cite{RIDNet}). Performance is evaluated in terms of PSNR/SSIM values. Our method can further elevate the performance of the state-of-the-art denoising methods.}
\label{table_quantitative_results}

\vspace{2Ex}

\centering
\begin{tabular}{|c|c|c|c|}
\hline
Method & Urban100 & DIV2K & BSD68 \\ \hline
DnCNN~\cite{DnCNN}    & 29.07 $\rightarrow$ 29.29 & 29.66 $\rightarrow$ 29.74 & 28.92 $\rightarrow$ 29.00\\
\hline
RDN~\cite{RDN}    & 29.36 $\rightarrow$ 29.56 & 29.84 $\rightarrow$ 29.91 & 29.02 $\rightarrow$ 29.10\\
\hline
RIDNet~\cite{RIDNet}    & 29.71 $\rightarrow$ 29.92 & 29.99 $\rightarrow$ 30.09 & 29.11 $\rightarrow$ 29.20\\
\hline
\end{tabular}
\vspace{1Ex}
\caption{Changes of PSNR values before and after fine-tuning DnCNN~\cite{DnCNN}, RDN~\cite{RDN}, and RIDNet~\cite{RIDNet} on different datasets. Although input images are corrupted by Gaussian random noise with $\sigma=40$, noise level is not given to \algoref{algorithm_proposed} during the fine-tuning phase.
Overall, the performance gaps are slightly reduced compared to the results in \tabref{table_quantitative_results}, but still show consistently better denoising results.}
\label{table_quantitative_results_blind}

\vspace{2Ex}

\centering
\begin{tabular}{|c|c|c|l|l|c|l|l|}
\hline
Method                        & Dataset           & \multicolumn{3}{c|}{SIDD+~\cite{SIDDplus}}    & \multicolumn{3}{c|}{Nam~\cite{Nam}}     \\ \hline
\multirow{5}{*}{MIRNet~\cite{MIRNet}} & Fully pre-trained & \multicolumn{3}{c|}{36.82 / 0.9160}        & \multicolumn{3}{c|}{38.17 / 0.9531}        \\ \cline{2-8} 
                        & $M=10$              & \multicolumn{3}{c|}{36.89 / 0.9181}        & \multicolumn{3}{c|}{38.29 / 0.9533}        \\ \cline{2-8} 
                        & $M=20$              & \multicolumn{3}{c|}{36.93 / 0.9193}        & \multicolumn{3}{c|}{38.36 / \textbf{0.9534}}        \\ \cline{2-8} 
                        & $M=40$              & \multicolumn{3}{c|}{\textbf{36.96 / 0.9211}}        & \multicolumn{3}{c|}{\textbf{38.43} / 0.9531}        \\ \hline
\multirow{5}{*}{RIDNet~\cite{RIDNet}} & Fully pre-trained & \multicolumn{3}{c|}{36.43 / 0.9138}        & \multicolumn{3}{c|}{39.08 / 0.9585}        \\ \cline{2-8} 
                        & $M=10$              & \multicolumn{3}{c|}{36.51 / 0.9166}        & \multicolumn{3}{c|}{39.16 / 0.9590}        \\ \cline{2-8} 
                        & $M=20$              & \multicolumn{3}{c|}{36.56 / 0.9190}        & \multicolumn{3}{c|}{39.22 / 0.9594}        \\ \cline{2-8} 
                        & $M=40$              & \multicolumn{3}{c|}{\textbf{36.61 / 0.9229}}        & \multicolumn{3}{c|}{\textbf{39.28 / 0.9599}}        \\ \hline
\multirow{5}{*}{CBDNet~\cite{CBDNet}} & Fully pre-trained & \multicolumn{3}{c|}{34.46 / 0.8580}        & \multicolumn{3}{c|}{36.51 / 0.9491}        \\ \cline{2-8} 
                        & $M=10$              & \multicolumn{3}{c|}{34.55 / 0.8625}        & \multicolumn{3}{c|}{36.65 / 0.9499}        \\ \cline{2-8} 
                        & $M=20$              & \multicolumn{3}{c|}{34.62 / 0.8664}        & \multicolumn{3}{c|}{36.77 / 0.9507}        \\ \cline{2-8} 
                        & $M=40$              & \multicolumn{3}{c|}{\textbf{34.75 / 0.8738}}        & \multicolumn{3}{c|}{\textbf{36.99 / 0.9522}}        \\ \hline
\end{tabular}
\vspace{1Ex}
\caption{Fine-tuning results of MIRNet~\cite{MIRNet}, RIDNet~\cite{RIDNet} and CBDNet~\cite{CBDNet} on the real noise datasets (SIDD+~\cite{SIDDplus} and Nam~\cite{Nam}). PSNR/SSIM values are measured. We can achieve better results with a larger $M$.}
\label{table_quantitative_results_real}

\vspace{2Ex}

\centering
\begin{tabular}{|c|c|c|c|c|c|c|}
\hline
Method     & N2S~\cite{Noise2self}   & \multicolumn{3}{c|}{ N2S~\cite{Noise2self} + ours}                                          \\ \hline
$\#$ updates & 500 iterations per image     & $M=200$           & $M=600$            & $M=1000$         \\ \hline
PSNR  & 29.40 & 29.87  & 31.01 & 31.29  \\ \hline
SSIM       & 0.638 & 0.728  & 0.768 & 0.791 \\ \hline
\end{tabular}
\vspace{1Ex}
\caption{Fine-tuning results on real noise dataset (SSID+). N2S~\cite{Noise2self} algorithm is used to train the baseline network (DnCNN~\cite{DnCNN}) and generate initially denoised results. The trained networks are updated by our method and can generate much better results in terms of PSNR and SSIM values.}
\label{table_quantitative_results_n2s}
\end{table*}

\begin{figure*}[]
\begin{center}
\includegraphics[width=1.0\linewidth]{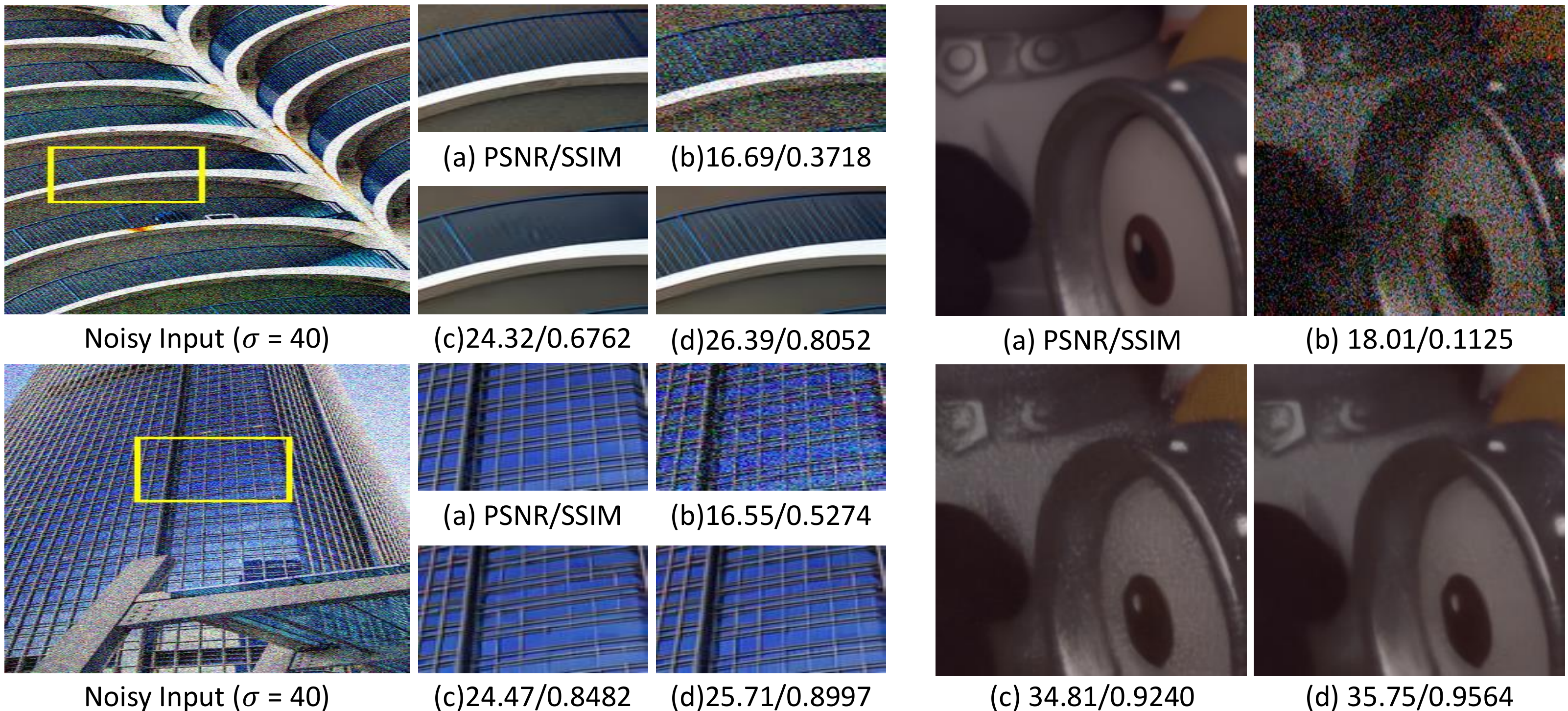}
\end{center}
\caption{Denoising results by RIDNet~\cite{RIDNet} and MIRNet~\cite{MIRNet}. We fine-tune for 40 iterations ($M=40$) to remove Gaussian noise (left) and real noise (right), and PSNR/SSIM values are compared. (a), (b), (c) and (d) denote the noisy-free, noisy, pre-trained and fine-tuned images respectively.}
\label{qual_results_rid}
\end{figure*}

\subsection{Gaussian noise denoising}
\paragraph{with known noise level}
In Table~\ref{table_quantitative_results}, we provide quantitative denoising results by removing Gaussian noise with known noise level.
DnCNN, RDN, and RIDNet are fine-tuned with the proposed learning method in \algoref{algorithm_proposed}. The results are measured on Urban100, BSD68 and DIV2K validation datasets.
Note that, the updated parameters of our denoisers can be well adapted to the specific inputs by increasing the number of iterations ($M$=10, 20, 40) and consistently produce better results on various Gaussian noise levels ($\sigma$=15, 25, 40). The performance gaps between the fully pre-trained baseline and fine-tuned networks become particularly large when the noise level is high (nearly 0.3dB on the Urban100 dataset with only 20 updates when $\sigma=40$).

\paragraph{with unknown noise level}
In Table~\ref{table_quantitative_results_blind}, we provide fine-tuning results by removing Gaussian denoising without knowing the noise level during the fine-tuning stage ($M$ = 40). Therefore, we use randomly generated noise level (between 0 and 50) to synthesize the Gaussian noise $\textbf{N}$ during the each fine-tuning iteration.
In comparison with the results in Table~\ref{table_quantitative_results}, 
the performance gain is slightly lower in this setting. However, the fine-tuned networks still consistently outperform the fully pre-trained baseline networks.

\subsection{Real noise denoising}
In Table~\ref{table_quantitative_results_real}, we show the fine-tuning results of RIDNet~\cite{RIDNet}, CBDNet~\cite{CBDNet}, and MIRNet~\cite{MIRNet} on the public datasets including real noise (SIDD+~\cite{SIDDplus} and Nam~\cite{Nam}). Similar to the Gaussian noise removal, our method further increases the denoising quality as the number of fine-tuning (i.e., $M$) goes up even with real noise.

Moreover, in \tabref{table_quantitative_results_n2s}, we compare with noise-to-self (N2S) which is the state-of-the-art self-supervision-based approach~\cite{Noise2self}.
We first denoise the real noisy images by using the DnCNN~\cite{DnCNN} pre-trained using N2S algorithm with the official code. Thereafter, we fine-tune the pre-trained DnCNN with our method. We observe that we need a relatively large number of iterations (several hundred times) to fine-tune the results from the initial parameter by N2S algorithm. However, the improved results demonstrate that our fine-tuning algorithm can be also embedded on top of self-supervised approaches as well as supervised methods.

\subsection{Visual results}
Please see our supplementary material. We provide extensive visual results on numerous settings, including synthetic Gaussian and real noise removal results.

In Figure~\ref{qual_results_rid}, we provide several denoising results. The input images are corrupted with high-level Gaussian noise ($\sigma$ = 40) and real noise.
RIDNet~\cite{RIDNet} and MIRNet~\cite{MIRNet} are fine-tuned with our algorithm for 40 iterations ($M$ = 40) to remove the synthetic and real noise respectively. Our fine-tuned denoiser can produce visually much better results and restore tiny details compared with the fully pre-trained state-of-the-art baseline networks. In particular, the regions with repetitive patterns (\eg., window frames) are restored surprisingly well, thereby verifying our spatially varying denoising performance.

\subsection{Run-time}
In a single update ($M$ = 1), our method takes approximately 0.035, 0.073, and 0.666 second to fine-tune the parameters of DnCNN, RIDNet, and MIRNet respectively with 256$\times$256 input image and Tesla V100 GPU.

\section{Conclusion}
In this work, we show that the performance of the conventional denoising methods can be improved by using the self-similar patches in the given input. Thus, we introduce a new and simple denoising approach that allows the update of the network parameters from the fully pre-trained version at test time, and enhance the denoising quality significantly by combining internal and large external information. Moreover, the proposed algorithm can be generally applicable to conventional denoising networks if they are fully convolutional without changing their original network architectures, and intensive and extensive experimental results demonstrate the superiority of the proposed method.

\FloatBarrier
{\small
\bibliographystyle{ieee_fullname}
\bibliography{cvpr}
}

\end{document}